\pdfoutput=1

\documentclass[11pt]{article}

\usepackage{emnlp2021}

\usepackage{times}
\usepackage{latexsym}

\usepackage[T1]{fontenc}

\usepackage[utf8]{inputenc}

\usepackage{microtype}

\usepackage{hyperref}
\usepackage{url}

\usepackage{graphicx}
\usepackage{amsfonts}
\usepackage{amsmath}
\usepackage{amssymb}
\usepackage{pifont}
\usepackage{bm}
\usepackage{nicefrac}
\usepackage{multirow}
\usepackage{subfigure}
\usepackage[font=small]{caption}
\usepackage{soul}
\usepackage{xcolor}
\usepackage{xspace}
\usepackage{booktabs}
\usepackage{comment}
\usepackage{sidecap}

\usepackage{tablefootnote}

\newcommand\rela{\text{ReLA}\xspace}
\newcommand\relu{\text{ReLU}\xspace}
\newcommand\att{\textsc{SMAtt}\xspace}
\newcommand\mhatt{\textsc{MHAtt}\xspace}
\newcommand\rmsnorm{\text{RMSNorm}\xspace}

\title{Sparse Attention with Linear Units}

\author{Biao Zhang$^1$ \quad Ivan Titov$^{1,2}$ \quad Rico Sennrich$^{3,1}$ \bigskip\\
  $^1$School of Informatics, University of Edinburgh \\
  $^2$ILLC, University of Amsterdam \\
  $^3$Department of Computational Linguistics, University of Zurich \\
  \resizebox{\textwidth}{!}{\texttt{B.Zhang@ed.ac.uk, ititov@inf.ed.ac.uk, sennrich@cl.uzh.ch}}
  }

\begin{document}
\maketitle
\begin{abstract}

Recently, it has been argued that encoder-decoder models can be made more interpretable by replacing the softmax function in the attention with its sparse variants.
In this work, we introduce a novel, simple method for achieving sparsity in attention: we replace the softmax activation with a \relu, and show that sparsity naturally emerges from such a formulation.
Training stability is achieved with layer normalization with either a specialized initialization or an additional gating function.
Our model, which we call Rectified Linear Attention (\rela), is easy to implement and more efficient than previously proposed sparse attention mechanisms.
We apply \rela to the Transformer and conduct experiments on five machine translation tasks.  \rela achieves  translation performance comparable to several strong baselines, with  training and decoding speed similar to that of the vanilla attention. Our analysis shows that \rela delivers high sparsity rate and head diversity, and the induced cross attention achieves better accuracy with respect to source-target word alignment than recent sparsified softmax-based models.
Intriguingly, \rela heads also learn to attend to nothing (i.e.\ `switch off') for some queries, which is not possible with sparsified softmax alternatives.\footnote{Source code is available at \url{https://github.com/bzhangGo/zero}.}

\end{abstract}

\section{Introduction}

\begin{figure}[t]
  \centering
  \mbox{
    \subfigure[\label{fig:vanilla_att} Attention with Softmax]{\includegraphics[scale=0.40]{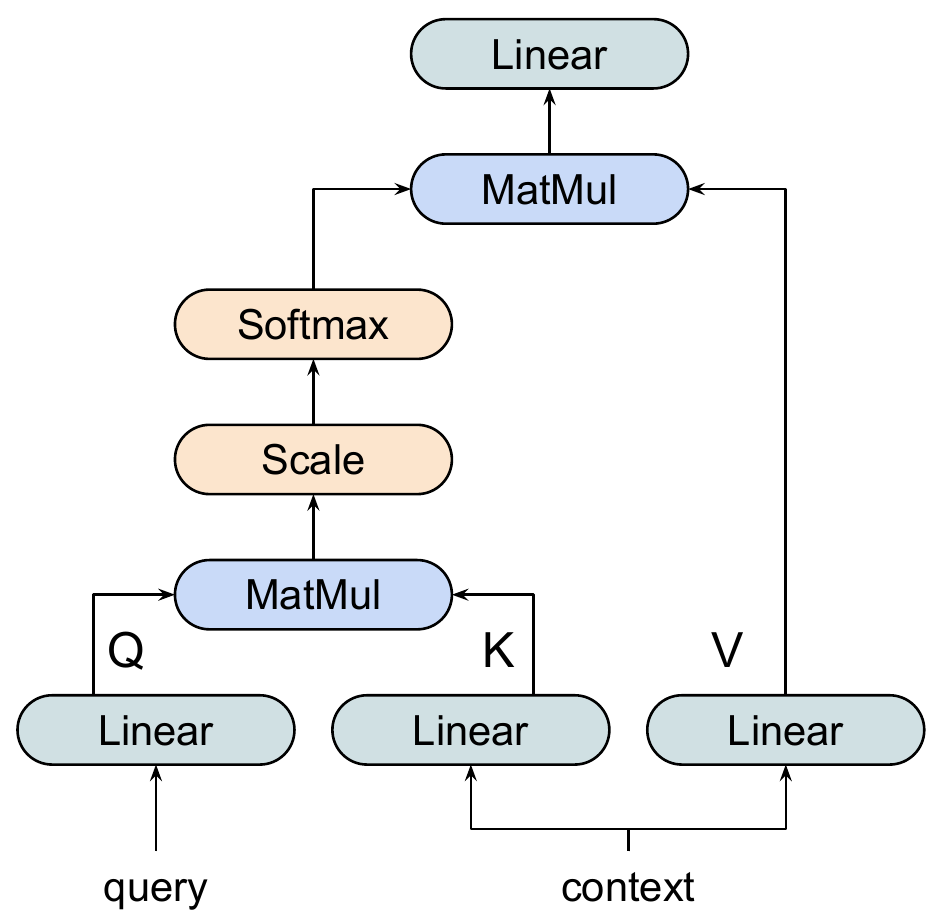}}
    \subfigure[\label{fig:rela} Rectified Linear Attention]{\includegraphics[scale=0.40]{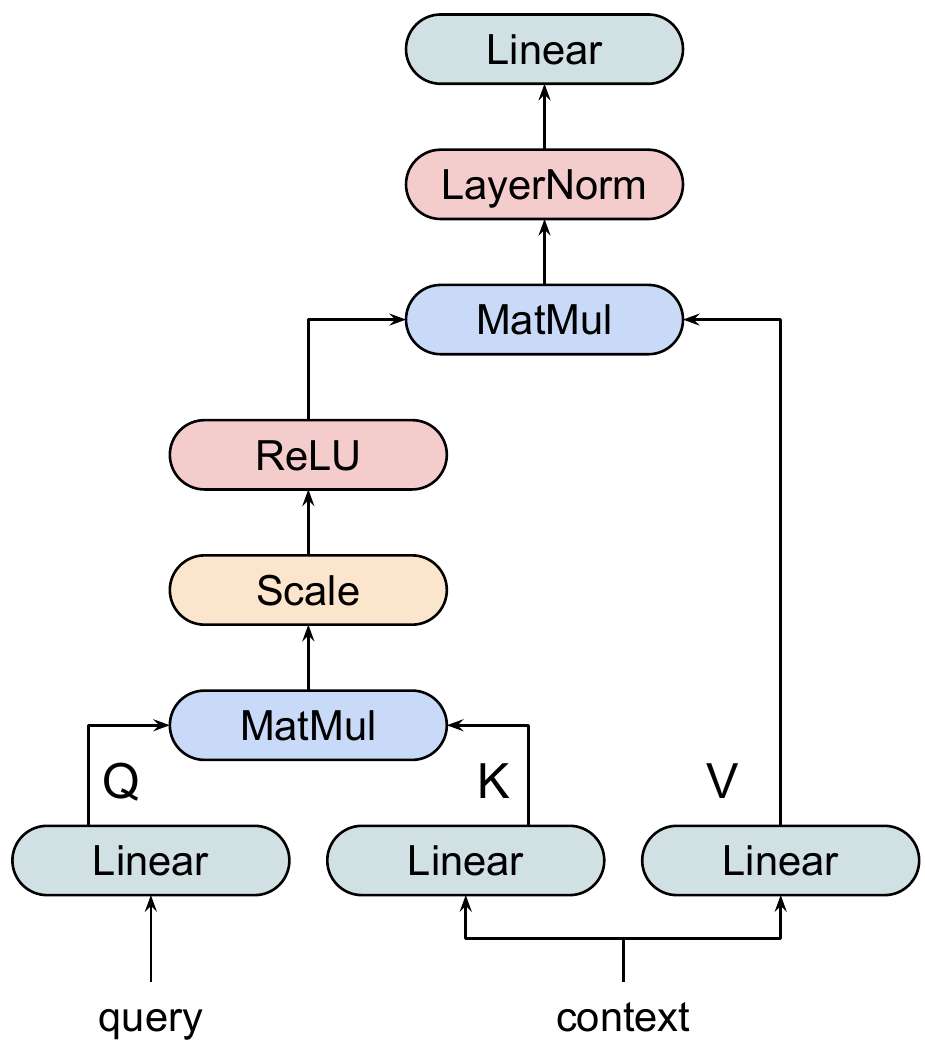}}
  }
  \caption{\label{fig:overview} Overview of the vanilla dot-product attention with softmax and the proposed Rectified Linear Attention (\rela). Major differences highlighted in red.}
\end{figure}

Attention  models~\cite{DBLP:journals/corr/BahdanauCB14} have been hugely successful recently, with Transformer~\cite{NIPS2017_7181_Transformer} in particular, advancing  state of the art  on various tasks, such as machine translation~\cite{bojar-etal-2018-findings}, document summarization~\cite{liu-lapata-2019-text} and speech processing~\cite{8462105}, and delivering a large impact on a broad range of NLP tasks via large-scale self-supervised pretraining~\cite{devlin-etal-2019-bert}. At the core of attention is a mechanism that dynamically highlights relevant context features for a given query input. In the vanilla softmax-based attention model~\cite[\att]{NIPS2017_7181_Transformer}, this is achieved by imposing a categorical distribution constraint on the query-context relevance (i.e. attention) scores, implemented with the softmax activation (see Figure \ref{fig:vanilla_att}).

\att produces dense distributions, assigning some small amounts of attention even to irrelevant features.
This complicates the analysis of the information flow in the model, and has led researchers to study sparse alternatives, which often lead to improved model performance and/or interpretability~\cite{correia-etal-2019-adaptively}. 
Efforts in this category include designing fixed sparsity patterns~\cite{raganato-etal-2020-fixed,child2019generating} and creating sparsified softmax variants~\cite{pmlr-v48-martins16,peters-etal-2019-sparse}. 
However, these methods also have drawbacks.
Fixed sparsity patterns lack flexibility and generalize poorly across tasks.
Sparsified softmax variants often depend on complex inference algorithms (e.g., requiring the sorting operation), which reduces their efficiency.

In this paper, we propose rectified linear attention (\rela) to alleviate the above problems.
\rela uses \relu rather than softmax as an activation function for attention scores, abandoning the probabilistic constraint.\footnote{Note that sparsified softmax variants also use some form of \relu to achieve sparsity, but they stick to the probabilistic constraint which demands extra complexity.}
\relu is inherently sparse since negative activations are dropped, and we will show that such sparse behaviour indeed emerges during training.
In contrast to softmax activations, the output of \relu can be any non-negative value, providing extra flexibility.
To stabilize gradients and ease model convergence, we apply layer normalization together with a specialized initialization or a gating mechanism. Figure \ref{fig:rela} shows \rela, and also contrasts it with \att. 

\rela is an easy-to-implement  drop-in replacement for \att that requires no specialized operations or inference processes.
Note that the behaviour of \rela is data-driven, and it does not enforce a constant attention mass or sparsity level across queries, even allowing for null attention (all attention scores are zero) for some queries.
We provide experimental results for \rela with Transformer on five machine translation tasks, along with an in-depth analysis on WMT14 English-German task. Our contributions are summarized below:
\begin{itemize}
    \item We propose \rela, a drop-in \att alternative, that learns sparse attention automatically with high flexibility and efficiency.
    \item Experiments on five translation tasks show that \rela achieves comparable translation performance, with similar training/decoding speed to \att, but is substantially faster than sparsified softmax baselines.
    \item Our analysis shows that \rela delivers high sparsity rate, high head diversity, and better accuracy than all baselines with respect to source-target word alignment. 
    We also observe the emergence of attention heads with a high rate of null attention, only activating for certain queries. For some heads, this null rate can also indicate the quality of sentence pairs.
\end{itemize}

\section{Related Work}

\rela ensures sparsity in attention. An alternative solution in this direction is to develop sparsified softmax alternatives, such as \textit{sparsemax}~\cite{pmlr-v48-martins16,malaviya-etal-2018-sparse}, \textit{entmax}~\cite{peters-etal-2019-sparse,correia-etal-2019-adaptively}, \textit{fusedmax}~\cite{NIPS2017_2d1b2a5f}, and hashing/clustering-based variants~\cite{roy2020efficient,kitaev2020reformer}. These models often require dedicated algorithms for forward and backward propagation, at the cost of a significant computational overhead. Another strategy is to manually define sparse patterns inspired by task-specific attention analysis. \citet{raganato-etal-2020-fixed} corroborated the feasibility of fixed patterns for Transformer encoder in translation. \citet{child2019generating} introduced local and strided patterns to scale \att up to very long inputs. Unlike data-driven approaches, whether these patterns could generalize to different tasks and settings is still an open question. 

In contrast, \rela is both data-driven and efficient. In this respect, our work shares similarity with the explicit sparse Transformer~\cite{zhao2019explicit} which also delivers faster speed but still depends on top-$k$ sorting as in {sparsemax} and {entmax} with $k$, a tunable hyperparameter. Note that all the above mentioned methods follow the categorical distribution constraint on attentions, while \rela goes beyond. Thus, unlike \rela, none of them enables null attentions.

A different type of linear attention model is proposed by
~\citet{katharopoulos20} and \citet{choromanski2020rethinking}, who aim at reducing the $\mathcal{O}(n^2)$ complexity in \att. These models behave fundamentally differently from \rela, because they eliminate the token-wise modeling rather than introducing sparsity.

The explanatory power of standard attention weights
is hotly debated~\cite{wiegreffe-pinter-2019-attention,jain-wallace-2019-attention}. Much of the criticism stems from the observation that low attention scores do not always imply irrelevance of the corresponding feature, as the information can still flow and its influence can be large (e.g., due to the large magnitude of the corresponding features). In contrast, sparse variants, including \rela, assign exact zeroes, ensuring that the information flow from the corresponding features within the attention component is cut completely. 
Even with standard attention, 
prior studies show some evidence that attention  partially  
reflects linguistic properties. In machine translation, the encoder-decoder attention captures the source-target word alignment to a certain degree~\cite{ghader-monz-2017-attention}, with recent work further strengthening this via specific induction methods~\cite{ding-etal-2019-saliency,kobayashi-etal-2020-attention,chen-etal-2020-accurate}. 
We apply analysis techniques from previous work to analyze our models.

\section{Background: Attention in Transformer}\label{sec:background}

Many variants of attention mechanism have been developed since its first proposal~\cite{DBLP:journals/corr/BahdanauCB14,luong-etal-2015-effective}. In this paper, we focus on the one used by Transformer, namely \textit{multi-head scaled dot-product attention} (\mhatt), in an encoder-decoder setup. 
Given query inputs $\mathbf{X} \in \mathbb{R}^{n\times d}$ and a sequence of context items $\mathbf{Y} \in \mathbb{R}^{m\times d}$, each head in \mhatt summarizes query-relevant context information as follows:
\begin{equation}\label{eq:dot_att}
    \begin{split}
        \text{\att} \left(\mathbf{X}, \mathbf{Y}\right) & = \bm{\alpha} \mathbf{V},  \\
        \text{with} \quad \bm{\alpha} &= \text{Softmax}\left(f\left(\mathbf{Q}, \mathbf{K}^T\right)\right),
    \end{split}
\end{equation}
with $\mathbf{Q} = \mathbf{X}\mathbf{W}_q; \mathbf{K}, \mathbf{V} = \mathbf{Y}\mathbf{W}_k, \mathbf{Y}\mathbf{W}_v$, where $n$ and $m$ are the query and context length, respectively; $d$ and $d_h$ are the model and head dimension, respectively; $\mathbf{W}_{*} \in \mathbb{R}^{d\times d_h}$ denotes trainable model parameters. 
$\bm{\alpha} \in \mathbb{R}^{n\times m}$ is the attention weight, which estimates the degree of relevance between one query input and each context. The softmax normalizes the scores and ensures that the attention weights $\bm{\alpha}$ define a categorical distribution. $f(\cdot)$ is a scoring function. Different attention mechanisms make different choices for $f(\cdot)$, but the use of softmax, or its sparsified variants, is universal.

\att in Transformer adopts the scaled dot product for $f(\cdot)$, which is further extended by \mhatt to allow for parallel attentions in different sub-spaces over the same inputs:
\begin{equation}\label{eq:mhatt}
    \text{\mhatt}\left(\mathbf{X}, \mathbf{Y}\right) = \left[\text{\att}^1, \ldots, \text{\att}^H\right] \mathbf{W}_o,
\end{equation}
where $[\cdot, \cdot]$ denotes the concatenation operation, $H$ is the number of heads, $\mathbf{W}_o \in \mathbb{R}^{Hd_h \times d}$ are  output transformation parameters, and $d=Hd_h$. 

In the encoder-decoder framework, \mhatt is used in three different ways:
\textit{Encoder Attention}, \textit{Decoder Attention} and \textit{Cross Attention}, modeling intra-source, intra-target, and source-target dependencies, respectively. 
Transformer performs layered \mhatt with residual connection and layer normalization~\cite{ba2016layer} to handle variations of token-wise dependencies. The learning of \mhatt is guided by the training objective, often without direct supervision.

\section{Rectified Linear Attention}

We argue that the use of the softmax function in \att (Eq.\ \ref{eq:dot_att}) has two undesirable consequences:
\begin{itemize}
    \item The attention mass is densely distributed over all context items, even  the ones that are intuitively irrelevant.
    \item The attention mass for each query is constant, although the relevance of context may vary.
\end{itemize}
Both potentially hamper interpretability and even performance.\footnote{As an anecdotal example, \citet{voita-etal-2018-context} performed an analysis of attention to previous sentences in MT, and found that the model has learned to generally attend to the end-of-sentence symbol as a way to ignore context. While this might be an effective strategy for instances where context matters little, this reduces the interpretability of attention.}

As an alternative to sparsified softmax variants~\cite{peters-etal-2019-sparse,correia-etal-2019-adaptively}, we go one step further and consider whether the softmax, or broadly the categorical distribution, could be avoided completely.

\paragraph{Model Structure} We offer an answer to the question by proposing rectified linear attention (\rela). \rela abandons the distribution assumption and adopts linear activation instead. It is formulated as follows (see Figure \ref{fig:rela} for illustration):
\begin{equation}\label{eq:rela}
    \begin{split}
        \text{\rela}\left(\mathbf{X}, \mathbf{Y}\right) & = \text{LN}\left(\bm{\alpha} \mathbf{V}\right),  \\
        \text{with} \quad \bm{\alpha} &= \text{\relu}\left(f\left(\mathbf{Q}, \mathbf{K}^T\right)\right),
    \end{split}
\end{equation}
where $f(\cdot)$ denotes any scoring function as in Eq. \ref{eq:dot_att}, $\text{LN}(\cdot)$ denotes variants of layer normalization~\cite{ba2016layer,NEURIPS2019_1e8a1942}, and $\text{\relu}(\cdot)=\max(0,\cdot)$ is the rectified linear unit. Note here, we describe our model by assuming only one attention head for clarity. In the multi-head \rela, we impose the normalization $\text{LN}(\cdot)$ on the concatenated head representation rather than each single head separately.

Unlike \att, \rela prunes out all negative scores of low query-context relevance, automatically ensuring the sparse property of the attention weight $\bm{\alpha} \in \mathbb{R}^{n\times m}$. 
Besides, \rela allows for null attention, where it assigns zero scores to all context items (i.e.\ some rows of $\bm{\alpha}$ are zero vectors), effectively switching off the corresponding attention head for certain queries.
Nevertheless, the outputs of \relu in Eq. \ref{eq:rela} are often of different scales and varied variance, causing gradient instability and also optimization failure. 

\paragraph{Stabilization with Normalization} A common strategy in deep learning to stabilize neuron activations is to apply layer normalization $\text{LN}(\cdot)$~\cite{ba2016layer}. We follow this strategy and normalize each representation $\mathbf{z} \in \mathbb{R}^{d_h}$ in the attention outputs ($\bm{\alpha}\mathbf{V}$) with root mean square layer normalization~\cite[\rmsnorm]{NEURIPS2019_1e8a1942}:
\begin{equation}\label{eq:rms_norm}
    \text{LN}(\mathbf{z}) = \text{\rmsnorm}(\mathbf{z}) = \frac{\mathbf{z}}{\text{RMS}(\mathbf{z})} \odot \mathbf{g},
\end{equation}
where $\odot$ denotes the element-wise multiplication, $\text{RMS}(\cdot)$ calculates the root mean square statistic, and $\mathbf{g} \in \mathbb{R}^{d_h}$ is the gain parameter, usually initialized at 1. 
We adopt \rmsnorm rather than the vanilla LayerNorm~\cite{ba2016layer} for \rela because it avoids the re-centering constraint, being more flexible and computationally simpler.

Although \rmsnorm largely smooths gradients, our preliminary experiments show that \rela still suffers from unstable gradients during early training, delivering suboptimal convergence. We propose two solutions, corresponding to two variants of \rela, to solve this problem by down-scaling \rela's activations.
\begin{description}
    \item[\rela-$i$] changes the initialization of the gain parameter $\mathbf{g}$ in \rmsnorm with a uniform xavier initializer: $\mathbf{g} \sim \mathcal{U}(-\sqrt{\frac{3}{d_h}}, \sqrt{\frac{3}{d_h}})$.\footnote{Note here we set $fan_{in}=fan_{out}=d_h$.} 
    \item[\rela-$g$] adds a simple gating function to the normalization:
    \begin{equation}
        \text{LN}(\mathbf{z}) = \sigma\left(\mathbf{w}\odot\mathbf{z}\right) \odot \text{\rmsnorm}(\mathbf{z}),
    \end{equation}
    where $\sigma(\cdot)$ denotes the sigmoid function, and $\mathbf{w} \in \mathbb{R}^{d_h}$ is a trainable parameter.
\end{description}
We compare their performance in our experiments. The only overhead due to \rela, compared to \att, is the added normalization layer and it is marginal. \rela is a drop-in replacement of \att, and we apply it to Transformer for all three types of attention. 

\section{Experiments}

\paragraph{Settings} We take machine translation as the testbed. We conduct experiments on five datasets of varied language pairs and training data sizes, including WMT14 English-German~\cite[En-De, 4.5M training instances]{bojar-EtAl:2014:W14-33}, WMT14 English-French~\cite[En-Fr, 36M]{bojar-EtAl:2014:W14-33}, WMT18 English-Finnish~\cite[En-Fi, 3.3M]{bojar-etal-2018-findings}, WMT18 Chinese-English~\cite[Zh-En, 25M]{bojar-etal-2018-findings}, and WMT16 Romanian-English~\cite[Ro-En, 608K]{bojar-etal-2016-findings}. 
We evaluate on the official test set from the corresponding year (e.g. newstest2014 for WMT14), and regard the previous year's test set as the development set (e.g. newstest2013 for WMT14). We preprocess all datasets using the byte pair encoding algorithm~\cite{sennrich-etal-2016-neural} with 32K merging operations.
We report detokenized case-sensitive BLEU~\cite{papineni-etal-2002-bleu} implemented by \textit{SacreBLEU}~\cite{post-2018-call},\footnote{Signature: BLEU+c.mixed+\#.1+s.exp+tok.13a+v.1.4.2} and also show tokenized case-sensitive BLEU with \textit{multi-bleu.perl} for ablation studies.

\begin{table}[t]
\centering
\small
\begin{tabular}{llc}
\toprule
{ID} & {Model} & BLEU \\
\midrule
1 & Baseline (softmax) & 26.9 (27.59) \\
2 & 1 + sparsemax & 26.4 (27.02) \\
3 & 1 + 1.5-entmax & 26.7 (27.39) \\
\midrule
4 & 1 + \relu alone & - \\
5 & 4 + \rmsnorm & 26.0 (26.60) \\
\midrule
6 & 1 + \rela-$i$ & 26.5 (27.16) \\
7 & 1 + \rela-$g$ & 26.6 (27.31) \\
\midrule
8 & 7 + LayerNorm & 26.6 (27.18) \\
9 & 7 + GeLU & 26.5 (27.14) \\
10 & 7 + Leaky \relu & 26.5 (27.13) \\
\midrule
11 & 7 + Encoder Attention Only & 26.3 (27.00) \\
12 & 7 + Decoder Attention Only & 27.0 (27.70) \\
13 & 7 + Cross Attention Only & 27.0 (27.69) \\
\bottomrule
\end{tabular}
\caption{\label{tb:wmt14_ende_ablation} SacreBLEU (tokenized BLEU in brackets) for different models and settings on WMT14 En-De. \textit{GeLU}: Gaussian error linear unit~\cite{hendrycks2016gaussian}; \textit{Leaky \relu}: leaky rectified linear unit~\cite{xu2015empirical}. \textit{Baseline}: Transformer. ``-'': optimization failed, where training loss didn't decrease. Higher BLEU indicates better result.}
\end{table}

\paragraph{Model Configuration} 
We use the Transformer base setting for experiments: model dimension $d=512$, head number $H=8$, head dimension $d_h=64$, 6 layers and FFN size of 2048~\cite{NIPS2017_7181_Transformer}. We apply dropout to the residual connections and attention weights, with a rate of 0.1. We tune model parameters using Adam~\cite[$\beta_1=0.9, \beta_2=0.98$]{kingma2014adam} with label smoothing of 0.1. We schedule the learning rate following~\citet{NIPS2017_7181_Transformer} with a warmup step of 4K. Each training batch contains around 25K target tokens. For decoding, we average the last 5 checkpoints and adopt beam search with a beam size of 4 and length penalty of 0.6.

Apart from softmax-based \att, we consider two additional baselines: \textit{sparsemax}~\cite{pmlr-v48-martins16} and \textit{1.5-entmax}~\cite{peters-etal-2019-sparse,correia-etal-2019-adaptively}.\footnote{\url{https://gist.github.com/justheuristic/60167e77a95221586be315ae527c3cbd}} 
We implement all models with Tensorflow (version 1.13.1).

\subsection{Translation Results}

We start with an ablation study for \rela on WMT14 En-De. Results are given in Table \ref{tb:wmt14_ende_ablation}.

\paragraph{Ablation on \rela's Architecture} At the heart of \rela is its replacement of softmax with \relu. But simply applying \relu to \att increases gradient instability, resulting in training failure (\textcircled{4}). Applying layer normalization to the outputs of the attention model alleviates this problem, albeit sacrificing 0.9 detokenized BLEU (\textcircled{1}$\rightarrow$\textcircled{5}). By contrast, the proposed solutions, \rela-$i$ and \rela-$g$, yield a detokenized BLEU score of 26.5 (\textcircled{6}) and 26.6 (\textcircled{7}) respectively, narrowing the quality gap against the baseline. \rela-$g$ performs slightly better than \rela-$i$ (+0.1 detokenized BLEU) and on par with 1.5-entmax (-0.1 detokenized BLEU), partially due to the increased gating parameters.  In the following experiments and analysis, we mainly report results with \rela-$g$ (i.e. \textcircled{7}).\footnote{Note we apply \rela-$g$ to all attention sublayers so as to avoid the interference of other attention variants. This allows us to fully examine the effectiveness of \rela.}

\begin{table}[t]
\centering
\small
\begin{tabular}{lccc}
\toprule
{Model} & {\#Params} & $\Delta$Train & $\Delta$Decode \\
\midrule
softmax & 72.31M & 1.00$\times$ & 1.00$\times$ \\
sparsemax & 72.31M & 0.26$\times$ & 0.54$\times$ \\ 
1.5-entmax & 72.31M & 0.27$\times$ & 0.49$\times$ \\
\midrule
\rela-$g$ & 72.34M & 0.93$\times$ & 0.98$\times$ \\
\bottomrule
\end{tabular}
\caption{\label{tb:wmt14_ende_efficiency} Number of parameters (\#Params) and running efficiency for different models on WMT14 En-De. \textit{$\Delta$Train}: speedup per training step measured on 500 steps with about 25K target tokens per batch. \textit{$\Delta$Decode}: translation speedup on newstest2014 with a batch size of 1. We perform 3 runs on a single GeForce GTX 1080 Ti and report average results for the speedups. Higher speedup indicates better efficiency.}
\end{table}

\begin{table*}[t]
\centering
\small
\begin{tabular}{lcccc}
\toprule
{Model} & WMT14 En-Fr & WMT18 En-Fi & WMT18 Zh-En & WMT16 Ro-En \\
\midrule
softmax & 37.2 & \textbf{15.5} & \textbf{21.1} & 32.7  \\
sparsemax & 37.3 & 15.1 & 19.2 & \textbf{33.5}  \\
1.5-entmax & \textbf{37.9} & \textbf{15.5} & 20.8 & 33.2 \\
\rela-$g$ & \textbf{37.9} & 15.4 & 20.8 & 32.9 \\
\bottomrule
\end{tabular}
\caption{\label{tb:more_mt_results} SacreBLEU scores for different models on more WMT translation tasks. Best scores are highlighted in \textbf{bold}.}
\end{table*}

\paragraph{\rmsnorm vs.\ LayerNorm} Results show that replacing \rmsnorm with LayerNorm (\textcircled{7}$\rightarrow$\textcircled{8}) leads to no quality improvement (-0.13 tokenized BLEU). We adopt \rmsnorm for \rela due to its efficiency. 

\paragraph{\relu vs.\ its Variants} We also attempted some smoothed variants of \relu, such as GeLU~\cite{hendrycks2016gaussian} and Leaky \relu~\cite{xu2015empirical}. Results show that these variants (\textcircled{9}, \textcircled{10}) yield worse performance than \relu (-0.1 detokenized BLEU). Dropping those low-relevance attention scores, as \rela does, benefits translation.

\paragraph{\rela for Different Attention Types} By default, we apply \rela to all attention sublayers. As shown in Section \ref{sec:background}, Transformer includes three types of attentions with different functionalities. We study next how \rela performs when applied to each attention separately. Results show that incorporating \rela into the decoder self-attention (\textcircled{12}) or encoder-decoder cross attention (\textcircled{13}) yields quality gains over Baseline (+0.1 detokenized BLEU). By contrast, only sparsifying encoder-side attentions with \rela leads to a big quality reduction (\mbox{-0.6} detokenized BLEU). We argue that the encoder self-attention requires denser token-wise modeling to induce informative features of the source input for translation, compared to the other two attention types, echoing with the findings of~\citet{correia-etal-2019-adaptively}.

\paragraph{Efficiency Analysis}

Table \ref{tb:wmt14_ende_efficiency} shows the results. Sparsemax and 1.5-entmax run more than 3 and 1.8 times slower than Baseline (softmax) at training and decoding, respectively. We ascribe this to the dedicated inference procedure (such as sorting) both methods require in order to discover the best sparsity patterns~\cite{peters-etal-2019-sparse}, which reduces efficiency. By contrast, the computation in \rela-$g$ is much simpler, and training and decoding speed is comparable to the baseline.

\begin{figure}[t]
  \centering
  \includegraphics[scale=0.50]{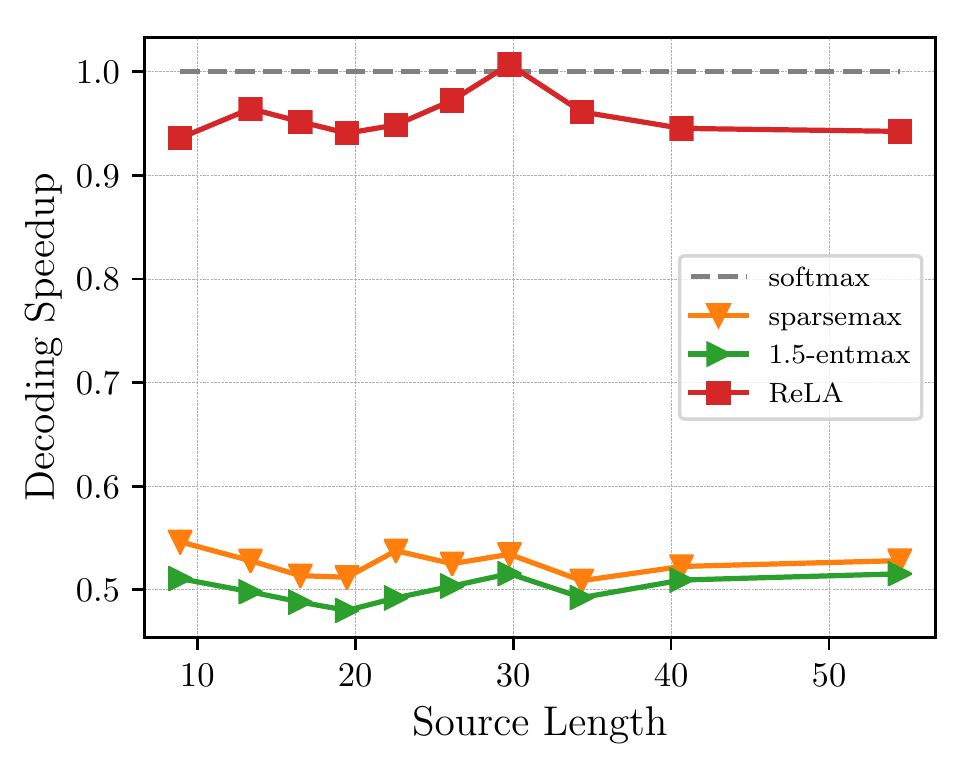}
  \caption{\label{fig:source_length} Decoding speedup as source length increases on WMT14 En-De. We divide the newstest2014 testset into 10 disjoint groups uniformly according to source length. Results are averaged over 3 runs.}
\end{figure}

\begin{figure*}[t]
  \centering
  \includegraphics[scale=0.60]{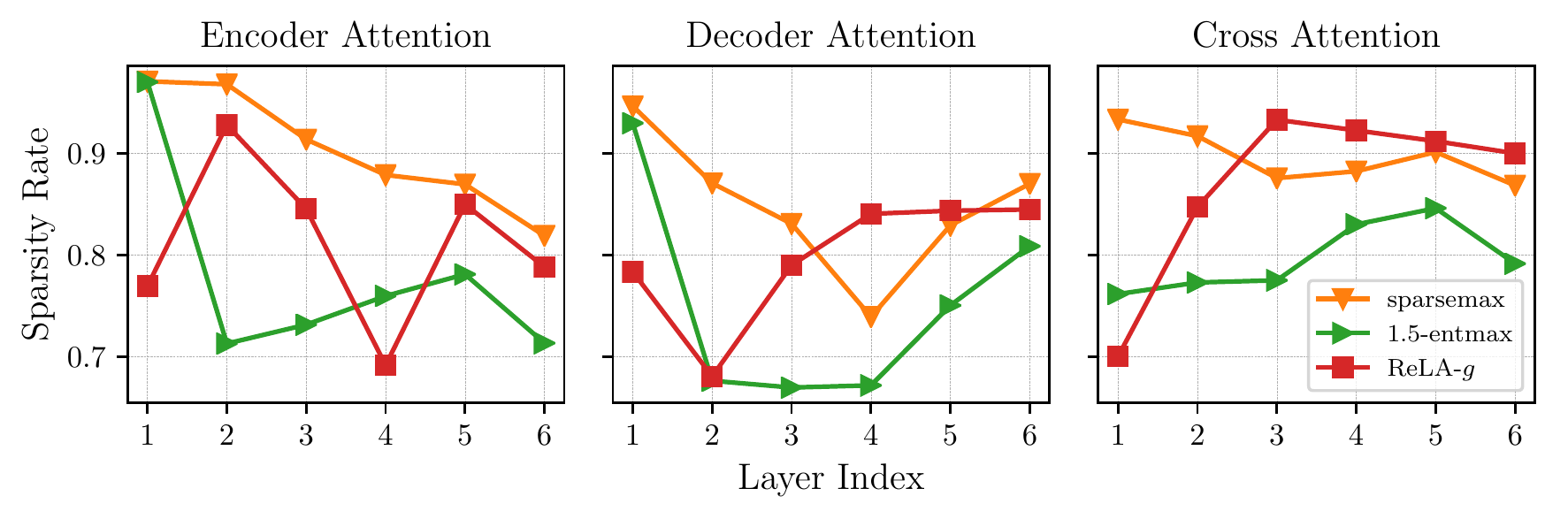}
  \caption{\label{fig:sparsity} Average sparsity rate over heads at each layer for different attention models and types on the WMT14 En-De test set. Larger sparsity rate indicates that more attention scores are exactly zero.}
\end{figure*}

Besides, we offer an analysis about the impact of source length on decoding speed. Curves in Figure \ref{fig:source_length} show consistent efficiency trend across different lengths: \rela translates slightly slower than Baseline but at least $1.8$ times faster than sparsemax and 1.5-entmax. 

We also notice that \citet{correia-etal-2019-adaptively} and \citet{zhao2019explicit} reported better training efficiency for sparsemax and 1.5-entmax than our results in Table \ref{tb:wmt14_ende_efficiency}. This is due to implementation difference. We re-tested the efficiency of different approaches using Pytorch with an in-house Transformer codebase, and worked with the official entmax implementation\footnote{Available at \url{https://github.com/deep-spin/entmax}.}. We observe that the training efficiency gap becomes much narrower, where sparsemax, 1.5-entmax and \rela yield a speedup of 0.87$\times$, 0.90$\times$ and 0.95$\times$, respectively. Although speedups vary across implementations, \rela shows consistently higher computational efficiency than these sparsified softmax variants.

\paragraph{Why \rela-$g$ Is Slower Than Softmax?}

\begin{table}[t]
\centering
\small
\begin{tabular}{lc}
\toprule
{Model} & FLOPs\\
\midrule
softmax & $3HT^2 - HT$ \\
\rela-$g$ & $HT^2 + 10Td + T$ \\
\bottomrule
\end{tabular}
\caption{\label{tb:flops} Comparison of FLOPs between softmax-based \att{} and \rela-$g$.}
\end{table}

Table \ref{tb:wmt14_ende_efficiency} and Figure \ref{fig:source_length} show that \rela-$g$ runs slower than Baseline. This is because \rela-$g$ is not just an activation function as in softmax. Apart from \relu, \rela-$g$ also includes a gated \rmsnorm  layer which brings in extra computational overhead. This becomes clearer as we show their FLOPs in Table~\ref{tb:flops}, where $T$ denotes the sequence length.

Take Transformer base ($H=8, d=512$) as an example. For translation tasks where sequences often contain fewer than 100 tokens, the FLOPs of softmax is lower than that of \rela-$g$ ($239K < 592K$, at $T=100$). But \rela-$g$ has better scalability with respect to sequence length and would benefit long-sequence modeling ($23.99M > 13.12M$, at $T=1000$).

\paragraph{Results for More Language Pairs}

Table \ref{tb:more_mt_results} summarizes the results.
Overall, performance of \rela-$g$ is competitive to the baseline, with BLEU differences ranging from -0.3 detokenized BLEU (Zh-En) to +0.7 detokenized BLEU (En-Fr), suggesting that \rela generalizes to different (similar or distant) language pairs. 
Average performance is 0.5 detokenized BLEU higher than that of sparsemax, and 0.1 detokenized BLEU below that of 1.5-entmax.

\subsection{Attention Analysis}

Although different sparsified \att models achieve comparable translation performance, their learned attention weights $\bm{\alpha}$ often have different characteristics. In this section, we quantitatively analyze these weights on WMT14 En-De. We first define \textbf{layer attention}, \textit{the weights averaged over the heads in one layer}, to ease our following study. 
Besides, we obtain the word-level attention weights by merging its subword-level counterparts following~\citet{zenkel2019adding}. We train each model three times with different random seeds on WMT14 En-De and report the average results. 

\paragraph{Attention Sparsity} The ability to automatically induce sparse attention is one of the key characteristics of \rela. We next report the sparsity rates, i.e. the fraction of attention weights exactly equal to 0.
We calculate the average sparsity rate over heads for each layer.

\begin{figure}[t]
  \centering
  \includegraphics[scale=0.60]{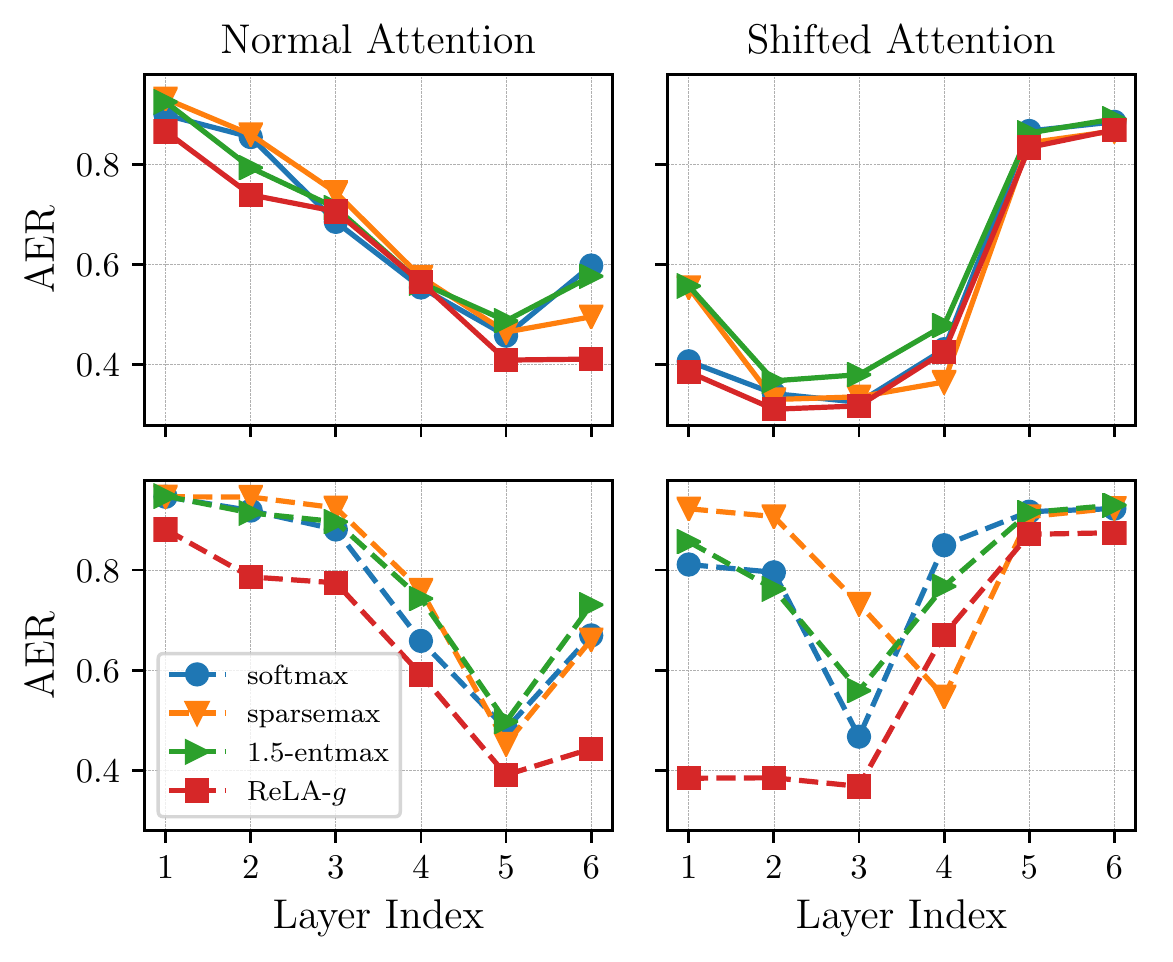}
  \caption{\label{fig:aer} AER scores for different models with normal attention (left, $\bm{\alpha}$) and shifted attention (right, $\bm{\alpha}[1:]$). Solid curves correspond to the best head per layer; dashed curves are for layer attention. Lower AER score indicates better alignment quality.}
\end{figure}

\begin{figure*}[t]
  \centering
  \includegraphics[scale=0.60]{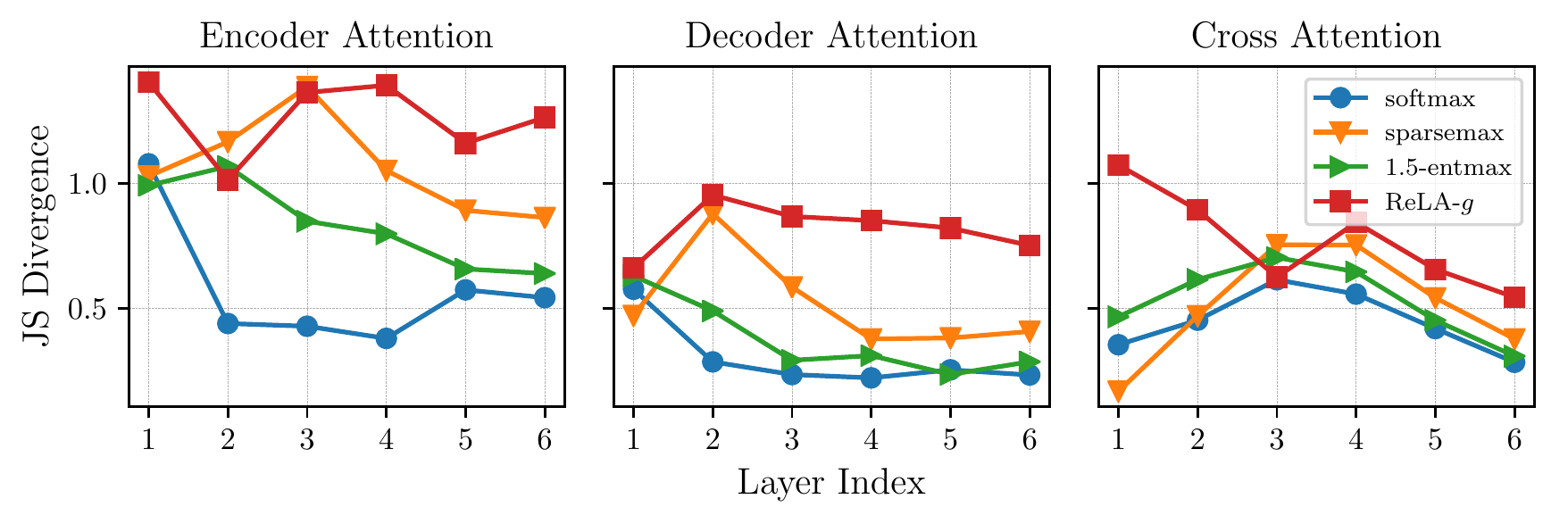}
  \caption{\label{fig:diversity} Jensen-Shannon (JS) Divergence over heads at each layer for different attention models and types on the WMT14 En-De test set. Higher JS Divergence indicates higher head diversity.}
\end{figure*}

Results are shown in Figure \ref{fig:sparsity}. We observe that the cross attention has the highest sparsity rate on average, resonating with the fact that word alignment is sparse. Self-attention at lower encoder/decoder layers often has a higher sparsity rate, particularly for sparsemax and 1.5-entmax. In \rela-$g$, we find that its sparsity rate for the decoder self- and cross-attention tends to increase with layer depth, while that of the encoder self-attention fluctuates. Overall, \rela-$g$ produces attentions of similar but slightly higher (lower) sparsity rate than 1.5-entmax (sparsemax), learned automatically without any constraint. Note softmax-based \att only produces dense attentions, i.e., a sparsity rate of 0.

\paragraph{Cross Attention vs. Word Alignment} We experiment with the publicly available De-En evaluation set\footnote{\url{https://www-i6.informatik.rwth-aachen.de/goldAlignment/}} and evaluate the alignment quality with alignment error rate~\cite[AER]{och-ney-2000-improved}. We study normal attention and shifted attention following previous work~\cite{chen-etal-2020-accurate,kobayashi-etal-2020-attention}. The former explores attention weights corresponding to decoder outputs (i.e. $\bm{\alpha}$ in Eq. \ref{eq:dot_att} and \ref{eq:rela}); the latter, by contrast, skips the weights at the first decoding step, i.e. $\bm{\alpha}[1:]$, to offset the left padding to the decoder inputs made for auto-regressive generation in Transformer.

Figure \ref{fig:aer} shows the results. Regardless of the attention type (normal or shifted), attention resembles alignments more at some middle layer of Transformer; and the shifted attention overall performs better than the normal attention, echoing previous findings~\cite{chen-etal-2020-accurate,kobayashi-etal-2020-attention}.
When considering the best AER head per layer, we observe that \rela-$g$ generally obtains the (near-)best AER at each layer index for both normal and shifted attention.
This becomes more obvious for the layer attention (bottom figures). Results in Table \ref{tb:aer} further show that the behaviour of \rela-$g$ is more alignment-like than the baselines we consider.

\begin{table}[t]
\centering
\small
\begin{tabular}{lcccc}
\toprule
\multirow{2}{*}{Model} & \multicolumn{2}{c}{Normal Attention} & \multicolumn{2}{c}{Shifted Attention} \\
 & AoL & AoH & AoL & AoH \\
\midrule
softmax & 75.95 & 67.51 & 79.38 & 54.31 \\
sparsemax & 78.17 & 67.88 & 82.32 & 54.95 \\
1.5-entmax & 78.82 & 67.69 & 79.84 & 58.98 \\
\rela-$g$ & \textbf{64.46} & \textbf{61.64} & \textbf{59.24} & \textbf{52.41} \\
\bottomrule
\end{tabular}
\caption{\label{tb:aer} Average AER scores over layers for different models. \textit{AoL}: average for layer attention; \textit{AoH}: average for the best head. Best scores are highlighted in \textbf{bold}.}
\end{table}




\paragraph{Head Diversity} We evaluate head diversity with a generalization of Jensen-Shannon divergence following~\cite{correia-etal-2019-adaptively} to reflect disagreements between heads. For \rela-$g$, we re-normalize its attention scores via softmax, and regard the null attention as a special one-hot distribution putting all probability mass to a dummy zero vector, i.e. entropy of 0. 

Figure \ref{fig:diversity} shows the results. We observe that the heads of the encoder self-attention exhibit much higher disagreement than those of the other two attention types for all sparsified attention models. Overall, heads in \rela-$g$ are in less agreement than with the sparsified softmax alternatives, in most cases across different attention types. This indicates that \rela-$g$ is capable of inducing heads with different roles~\cite{voita-etal-2019-analyzing}.

\begin{table}[t]
\centering
\small
\begin{tabular}{lccc}
\toprule
{Model} & Enc & Dec & Cross \\
\midrule
softmax & 0.56 & 0.31 & 0.45 \\
1.5-entmax &  0.84 & 0.39 & 0.52 \\
\midrule
\rela-$g$ ($\tau$=1.00) & 1.26 & 0.81 & 0.78 \\
\rela-$g$ ($\tau$=0.50) & 0.97 & 0.65 & 0.69 \\
\rela-$g$ ($\tau$=0.25) & 0.92 & 0.65 & 0.71 \\
\rela-$g$ ($\tau$=0.10) & 0.92 & 0.66 & 0.72 \\
\bottomrule
\end{tabular}
\caption{\label{tb:diversity} Average head diversity score over layers for different models. \textit{Enc}: encoder attention; \textit{dec}: decoder attention; \textit{cross}: cross attention. \textit{$\tau$}: temperature used for the re-normalization.}
\end{table}

Note we convert the attention scores of \rela-$g$ into categorical distribution via softmax for diversity evaluation. Such re-normalization might have a large impact on the final diversity results. We next explore this impact by adding some temperatures ($\tau$) to $\bm{\alpha}$, i.e. $\bm{\alpha}^\tau$ (in Eq. \ref{eq:rela}) before applying softmax. Smaller temperatures will enforce smoothness into the output distribution, so alleviating the emergence of peaked distributions. Table \ref{tb:diversity} shows the results. The temperature indeed affects the diversity results but does not eliminate the diversity gap, and the diversity of ReLA gradually converges as $\tau$ goes smaller.

\begin{figure}[t]
  \centering
  \includegraphics[scale=0.50]{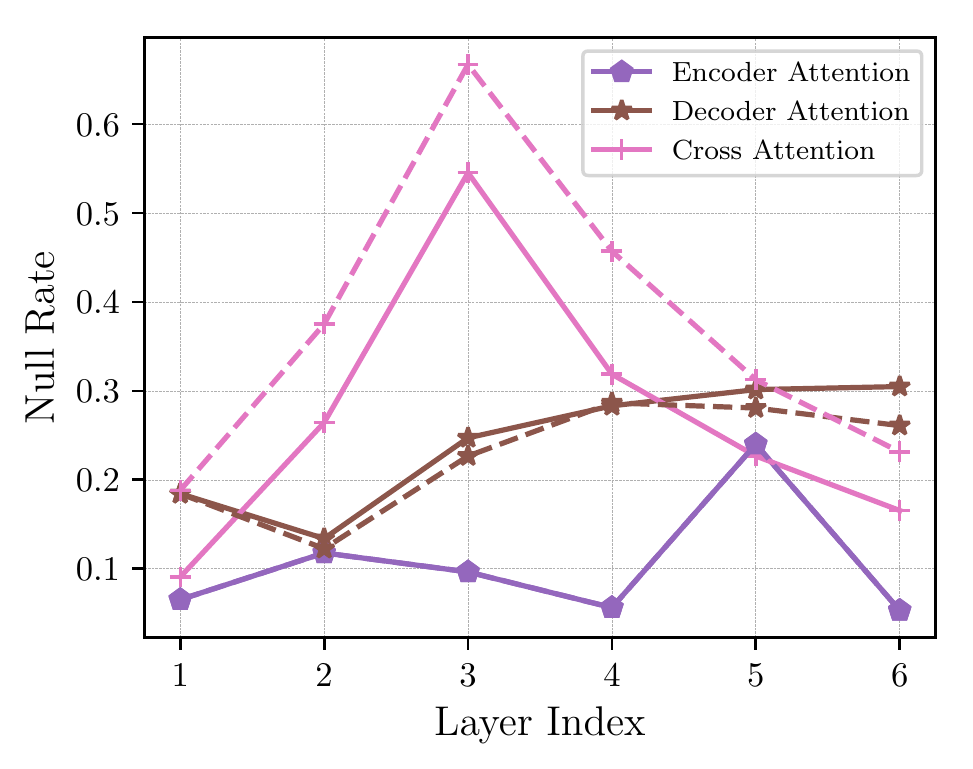}
  \caption{\label{fig:devitalization} Average null rate over heads at each layer for different attention types in \rela-$g$ on the WMT14 En-De test set. Null attention corresponds to attention of all zero scores. Dashed curves stand for the results on the hallucinated test set where target sentences are randomly shuffled. Hallucination pairs are assigned with significantly higher null rate for the cross-attention across different layers.}
\end{figure}

\begin{figure*}[t]
  \centering
  \includegraphics[scale=0.40]{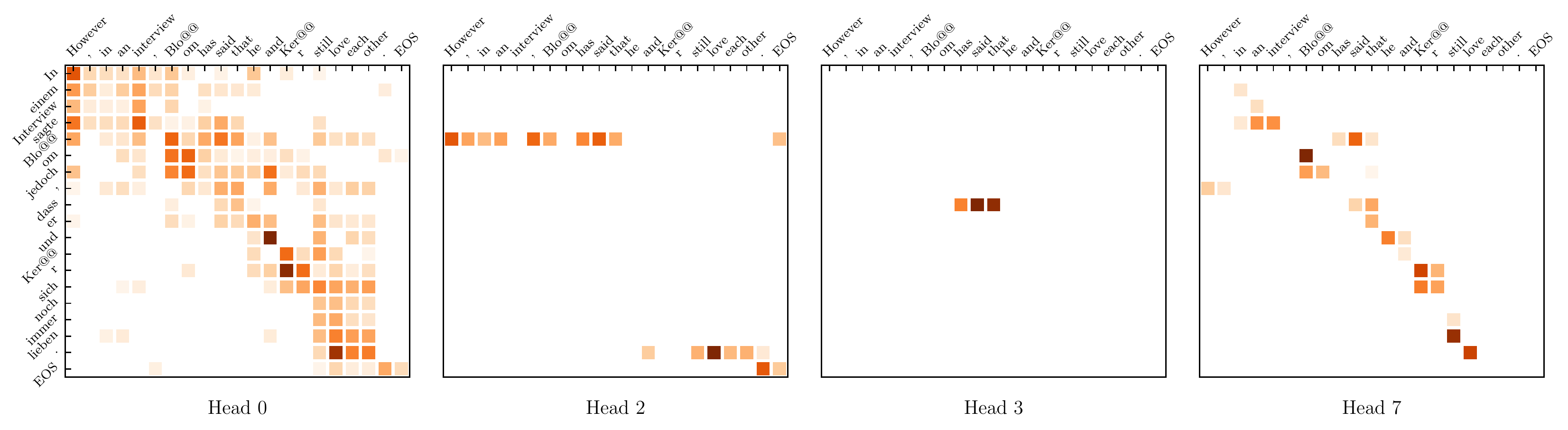}
  \caption{\label{fig:null_cases} Null-attention examples for head (0,2,3,7) at the 3rd cross attention layer. This example comes from the WMT14 En-De test set. Different heads show different linguistic patterns.}
\end{figure*}

\paragraph{Null Attention} One important feature distinguishing \rela from (sparsified) softmax is that \rela allows for null attention where all keys are assigned an attention weight of 0, effectively deactivating the attention head for this query. Figure \ref{fig:devitalization} analyzes the null rate of different attention types, i.e.\ the fraction of query tokens associated with all zero attention scores. Note all softmax-based variants have a null rate of about or exactly 0.

We find that the encoder self-attention has few null attentions, suggesting that encoder prefers denser connections and also compact information flow. The decoder self-attention yields more null attentions for deeper layers. Together with the findings from Figure \ref{fig:sparsity}, it shows that the lower decoder self-attention layers model dense dependency with previous target tokens, while the dependencies in the upper ones become sparser. The cross-attention shows the most interesting phenomenon: an obvious peak at the middle layer. Attention at this layer shows the highest sparsity (Figure \ref{fig:sparsity}) with a large null rate variance of 0.256 (over heads), low head disagreement (Figure \ref{fig:diversity}), but best AER score (right, bottom figure in Figure \ref{fig:aer}). Notice that attentions in \rela-$g$ are of high diversity. The layer attention at each layer has almost no null attentions.

Diving deeper into these null attentions as shown in Figure \ref{fig:null_cases}, we observe diverse specializations: head 0 and 7 capture source-target alignments with varying degrees of sparsity;
head 2 has a null rate of 83\%, and tends to fire after producing a verb (null rate 0\% after AUX, 23\% after VERB), attending tokens in the corresponding clause; head 3 has a null rate of 95\%, but regularly fires after comma (null rate 0.03\%), attending to the relevant source context (the clause boundary \textit{has said that} in this example). Extra attention matrices are shown in Appendix~\ref{app:null_att_case}.

\paragraph{Is Null Attention Meaningful?}
Apart from heads that have learned some sparse specialization, we also find that null attention can be informative for some cross-attention heads which learn an alignment. Specifically, the null rate increases for sentence pairs of low quality where many target tokens lack relevant source translations (see Appendix~\ref{app:null_att_case_ranking}).
In order to quantify this effect, we create a hallucinated test set with target references randomly shuffled for comparison. The dashed curves in Figure \ref{fig:devitalization} show that \rela-$g$ associates such hallucination pairs with clearly higher null rate for the cross-attention across different layers.

We next average the null rate of the cross-attention over layers and utilize this metric to rank the WMT14 En-De training corpus (top ranked samples have lower null rate). We randomly sample 100 cases from the top 10K candidates, and another 100 from the bottom 10K for manual analysis. We observe clear quality difference between these two groups: sentence pairs with a low null rate are predominantly good translations ($\sim$95\% correct), whereas sentence pairs with a high null rate are predominantly mistranslations ($\sim$1\% correct). Bad translations include sentence pairs with the wrong output language and semantically mismatched sentence pairs. Most interestingly, this null rate metric is sensitive to insertion errors, which are difficult to detect via traditional corpus filtering methods.
We note previous work that used attention statistics to identify mistranslations \cite{Rikters-Fishel2017MTSummit}, but find null attention more easily interpretable than more subtle changes in attention distribution.

\section{Conclusion and Future Work}

In this paper, we have presented rectified linear attention (\rela), a novel softmax-free sparse attention model. \rela avoids the categorical distribution assumption for attention, and, due to using \relu as activation function, prunes out all negative attention scores and produces sparse attention. To stabilize model training, we add a normalization layer to attention outputs with a specialized initialization or  gating structure. \rela is data-driven, computationally efficient and is a drop-in replacement of \att. 
Experiments on five machine translation tasks with Transformer demonstrate \rela's effectiveness in delivering comparable translation quality to softmax-based baselines. Results also show that \rela is substantially faster than sparsemax and 1.5-entmax at training and decoding. The attentions learned by \rela correspond better to word alignment, with high head diversity and sparsity rate.
Also, in contrast to alternative sparse attention approaches, \rela  produces null attentions, i.e.\ a head can assign a total attention of zero for some queries, leading to highly specialized attention heads and showing potential to indicate translation quality.

In the future, we will apply \rela to other neural models and tasks. 
We are interested in scaling \rela to very long inputs~\cite{child2019generating,kitaev2020reformer},
or multi-source architectures where the relevance of each source may vary.
In its current formulation, the sparsity level of \rela emerges from the threshold in \relu which prunes negative scores.
We are interested in ways to manipulate the level of sparsity,
or make the threshold differentiable.

\section*{Acknowledgments}

We thank the reviewers for their insightful feedback.
Rico Sennrich acknowledges support of the Swiss National Science Foundation (MUTAMUR; no. 176727).

\bibliography{naacl2021}
\bibliographystyle{acl_natbib}


\appendix



\section{Examples for Null Attention}\label{app:null_att_case}

\begin{figure*}[t]
  \centering
  \includegraphics[scale=0.40]{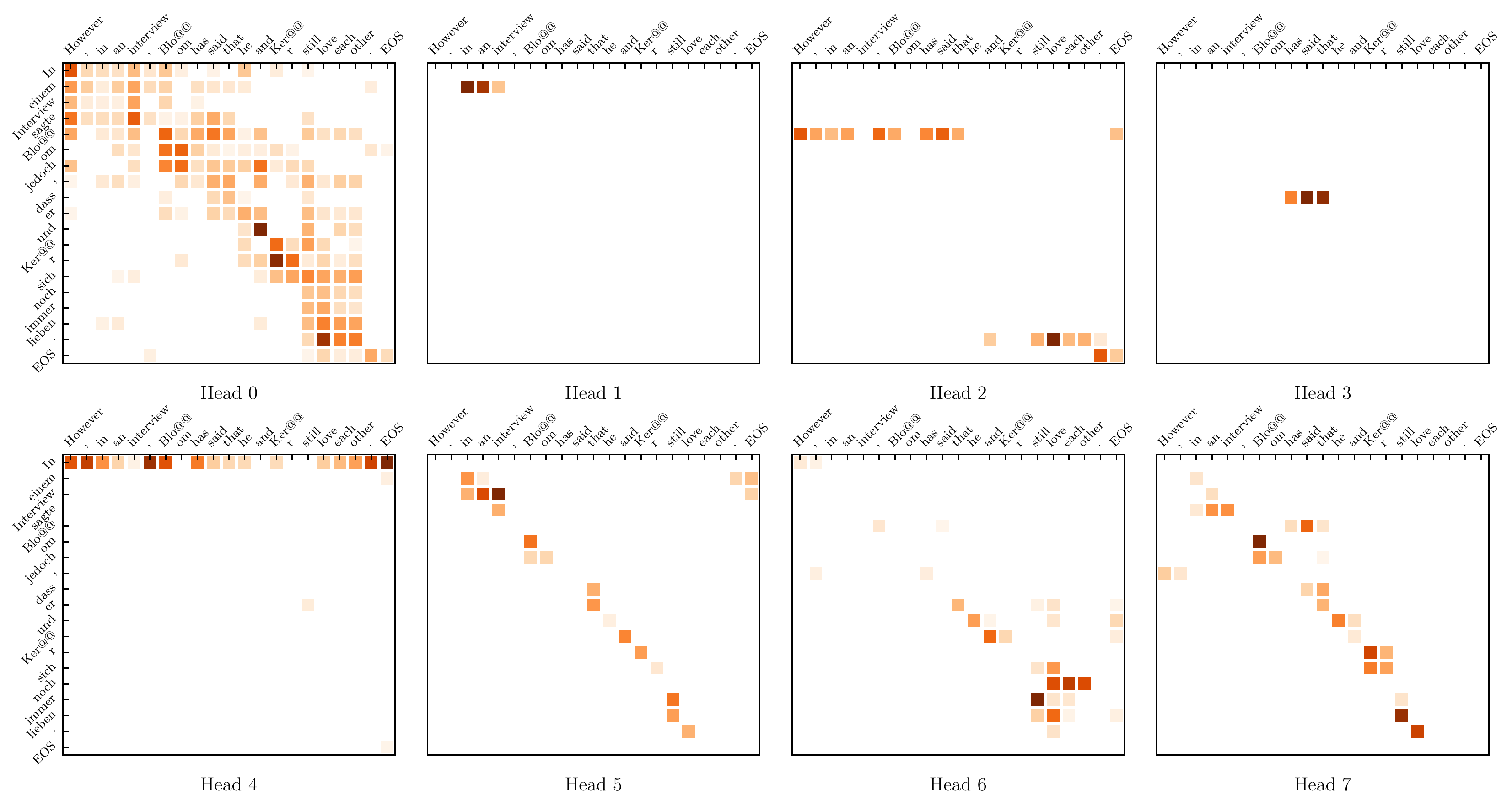}
  \includegraphics[scale=0.40]{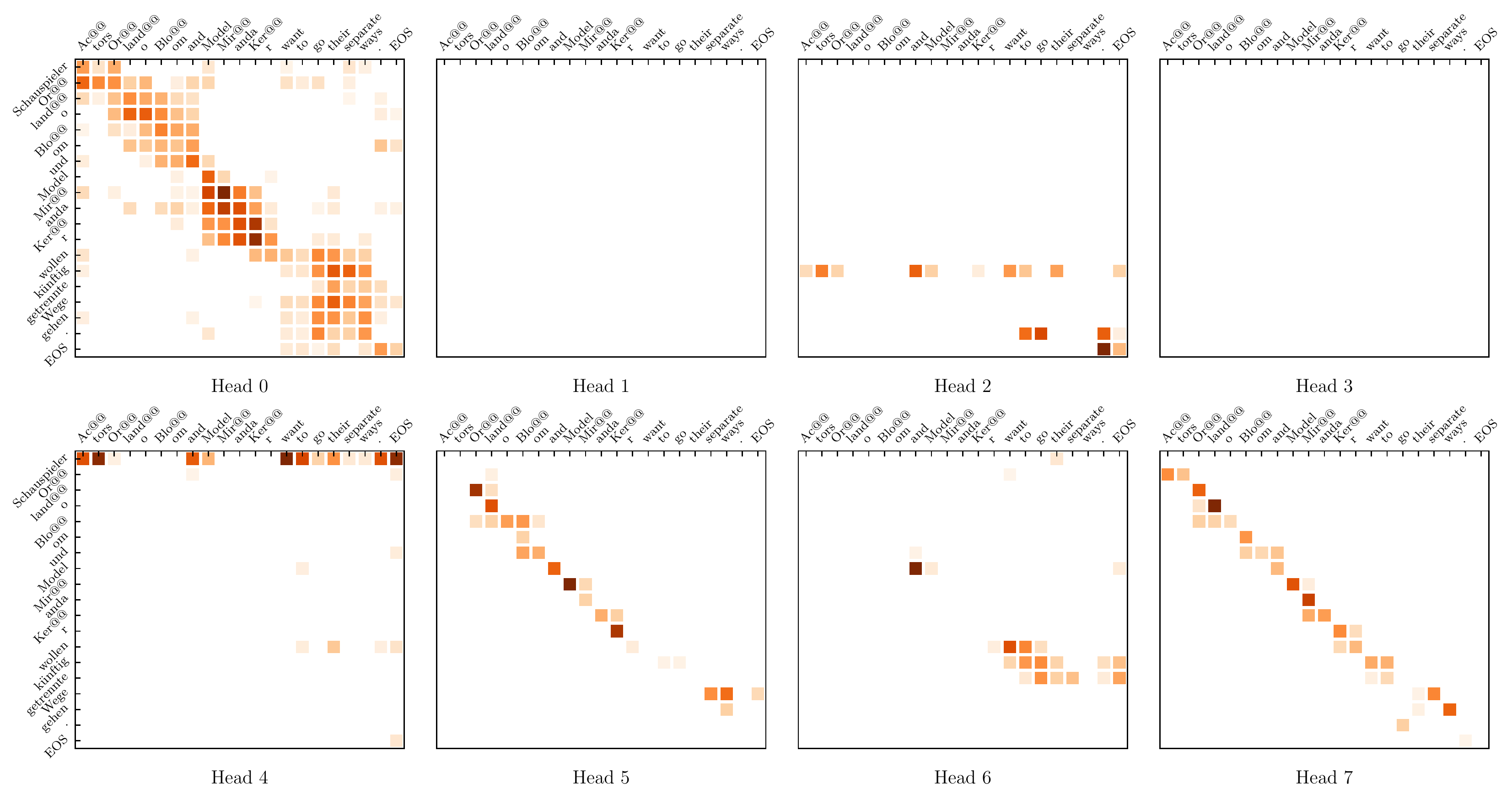}
  \includegraphics[scale=0.40]{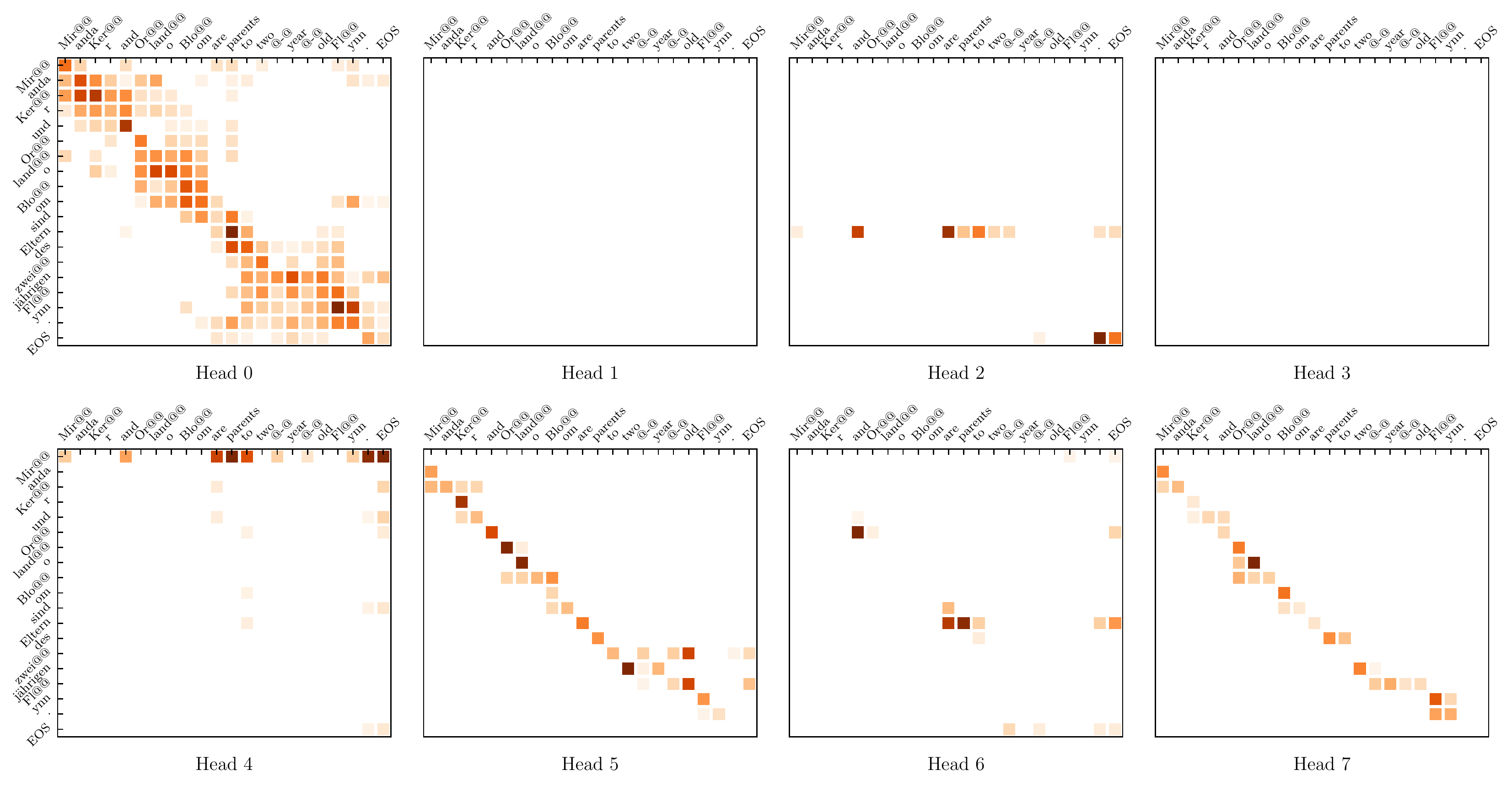}
  \caption{\label{fig:full_null_cases} Null-attention examples at the 3rd cross attention layer. All examples come from the WMT14 En-De test set.}
\end{figure*}
\begin{figure*}[t]
  \centering
  \includegraphics[scale=0.40]{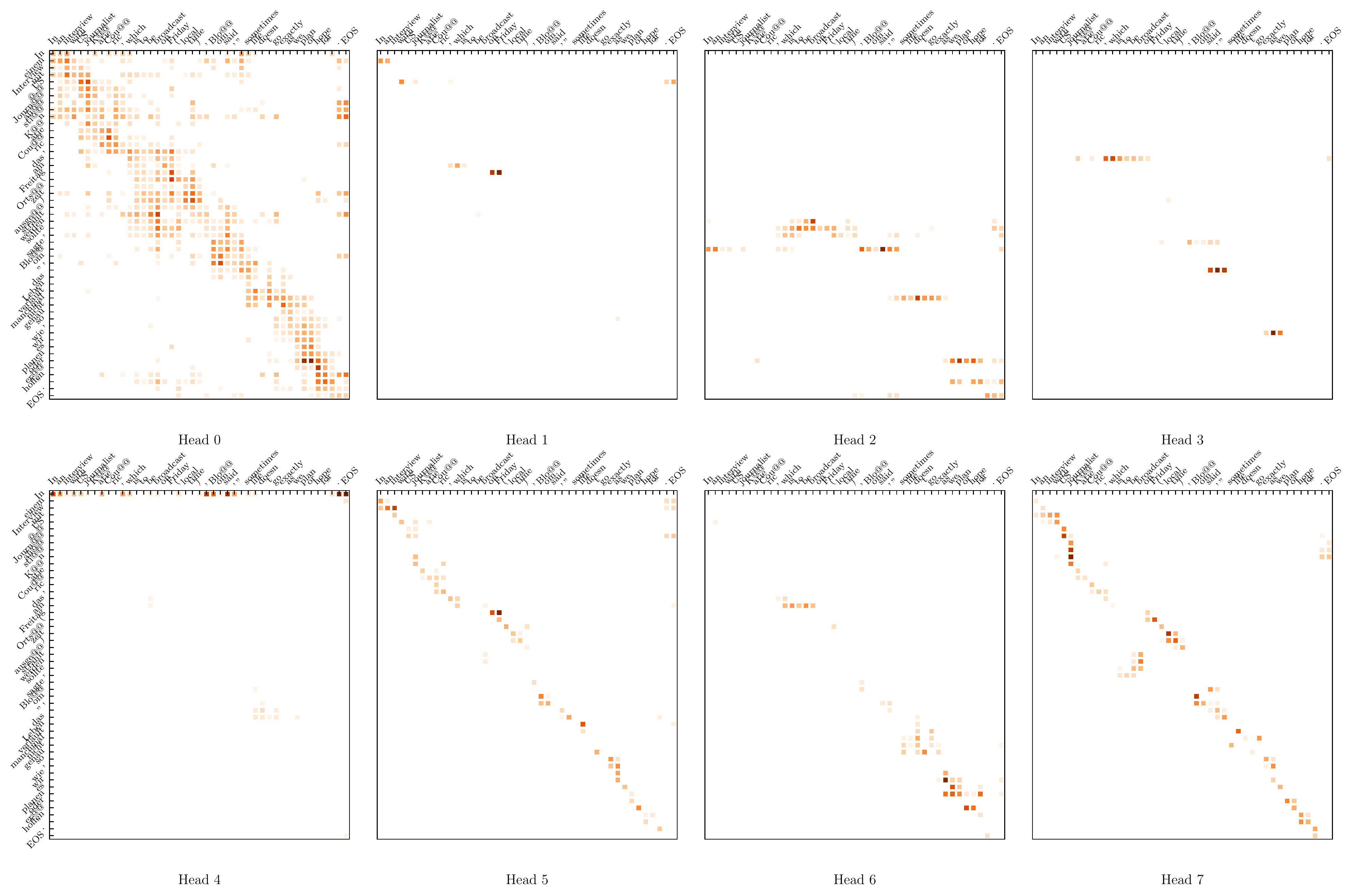}
  \caption{\label{fig:more_full_null_cases} Null-attention examples for multi-clause sentences at the 3rd cross attention layer. All examples come from the WMT14 En-De test set.}
\end{figure*}

\section{Null Attention for Diverse-quality Examples}\label{app:null_att_case_ranking}

\begin{figure*}[t]
  \centering
  \includegraphics[scale=0.50]{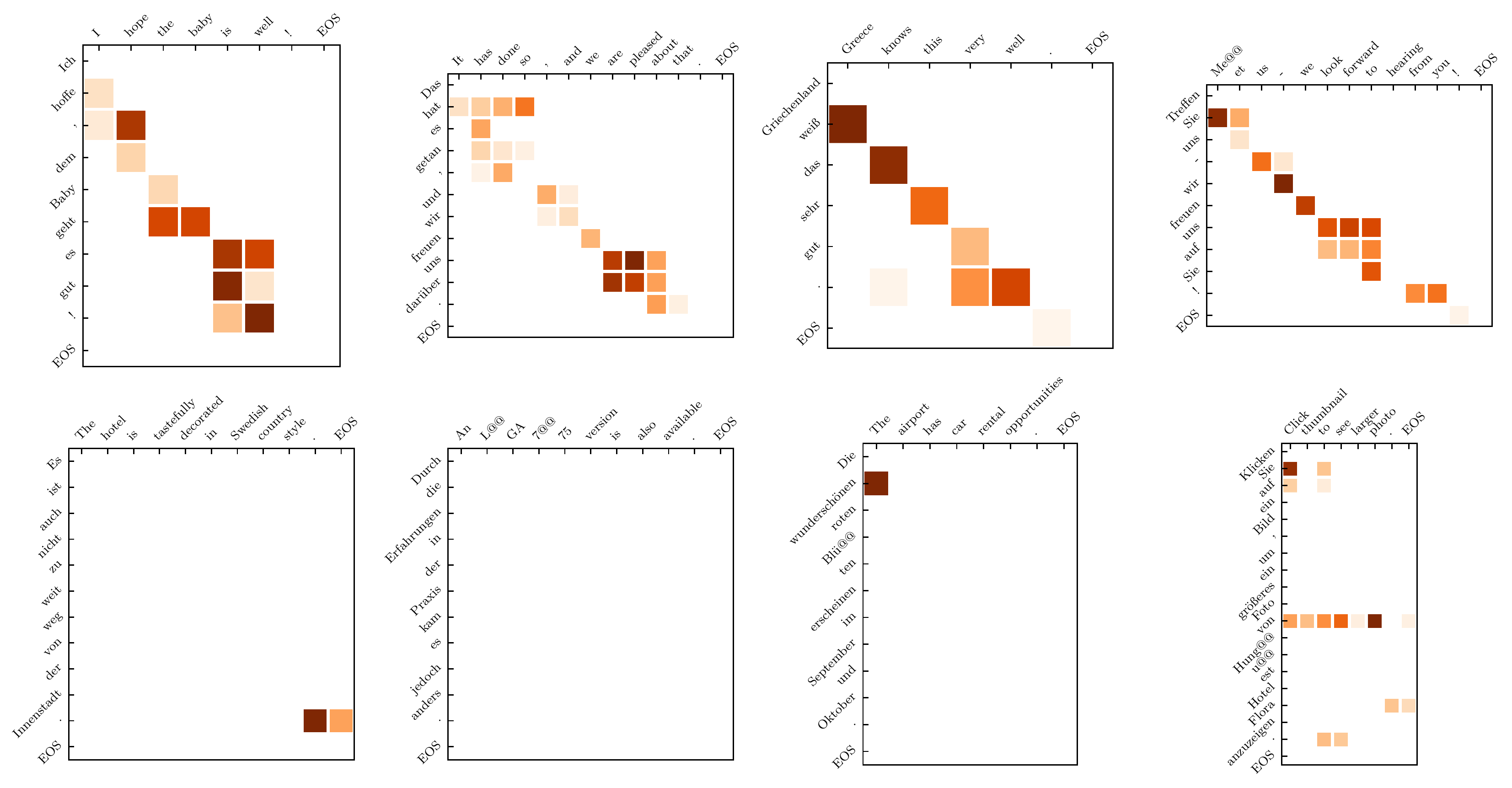}
  \caption{\label{fig:null_cases_ranking} Null-attention at the 3rd cross attention layer for the $7$-th head. Example comes from the WMT14 En-De training corpus. Top row shows high-quality examples, while bottom row shows low-quality ones. Low-quality examples include insertion errors and mistranslations, which increase the null rate.}
\end{figure*}

\end{document}